\documentclass[11pt]{article}

\usepackage[final]{acl}
\usepackage{times}
\usepackage{latexsym}
\usepackage[T1]{fontenc}
\usepackage[utf8]{inputenc}
\usepackage{microtype}
\usepackage{inconsolata}
\usepackage{graphicx}
\usepackage{amsmath}
\usepackage{amssymb}
\usepackage{booktabs}
\usepackage{multirow}
\usepackage{algorithm}
\usepackage{algpseudocode}
\usepackage{makecell}

\title{HiCAP: Hierarchical Cultural Adaptation via Personality for Dialogue Generation}

\author{
  \href{mailto:l.shen@soton.ac.uk}
  {\textbf{Lingzhi Shen\textsuperscript{1}}},
  \href{mailto:x.cai@soton.ac.uk}
  {\textbf{Xiaohao Cai\textsuperscript{1}}},
  \href{mailto:yunfei.long@qmul.ac.uk}
  {\textbf{Yunfei Long\textsuperscript{2}}},
  \href{mailto:imran.razzak@mbzuai.ac.ae}
  {\textbf{Imran Razzak\textsuperscript{3}}},
  \href{mailto:M.S.Jameel@southampton.ac.uk}
  {\textbf{Shoaib Jameel\textsuperscript{1}}}
\\
\\
  \textsuperscript{1}School of Electronics and Computer Science, University of Southampton, \\
  Southampton, United Kingdom
\\
  \textsuperscript{2}School of Electronic Engineering and Computer Science, Queen Mary University of London, \\
  London, United Kingdom
\\
  \textsuperscript{3}Department of Computational Biology, Mohamed bin Zayed University \\
of Artificial Intelligence, Abu Dhabi, United Arab Emirates
}

\begin{document}
\maketitle
\begin{abstract}
Cultural adaptation in dialogue systems refers to the ability to generate responses that conform to linguistic norms and social expectations across different cultures. However, existing dialogue generation approaches pay limited attention to the role of personality in shaping cultural expression and communication style. We therefore propose HiCAP, a novel framework that integrates personality traits into culturally adaptive dialogue generation. HiCAP employs a personality-conditioned control mechanism that hierarchically modulates response style by first aligning it with the cultural tone and then refining linguistic expression and content according to individual personality attributes. To embed these factors, the model combines supervised fine-tuning and instruction-based tuning, supported by dynamic instruction templates and multi-agent demonstrations to capture diverse combinations of culture and personality. Experiments on CulturalPersonas and CPED show that HiCAP achieves the strongest overall performance across various dialogue generation metrics, demonstrating effective cultural adaptability and personality modelling.
\end{abstract}

\section{Introduction}
The ability to understand and adapt to cultural context is a key step towards building human-centric artificial intelligence (AI) \cite{shen2026calm}. In conversational agents, cultural adaptation \cite{anik2025preserving} refers to the ability to flexibly adjust language and interaction strategies in accordance with the values, communication styles, and expectations associated with diverse cultural backgrounds, such as different countries, regions, or social groups. Such adaptability is essential for fostering trust and effective communication across global user populations. Accordingly, culturally aware dialogue agents have broad applications \cite{liu2025culturally}, including international customer service, multilingual virtual assistants, cross-cultural healthcare support, global education platforms, and mental health counselling, where contextually appropriate and respectful interactions are particularly important for minority groups.

\begin{figure}[tbp]
    \centering
    \includegraphics[width=\columnwidth]{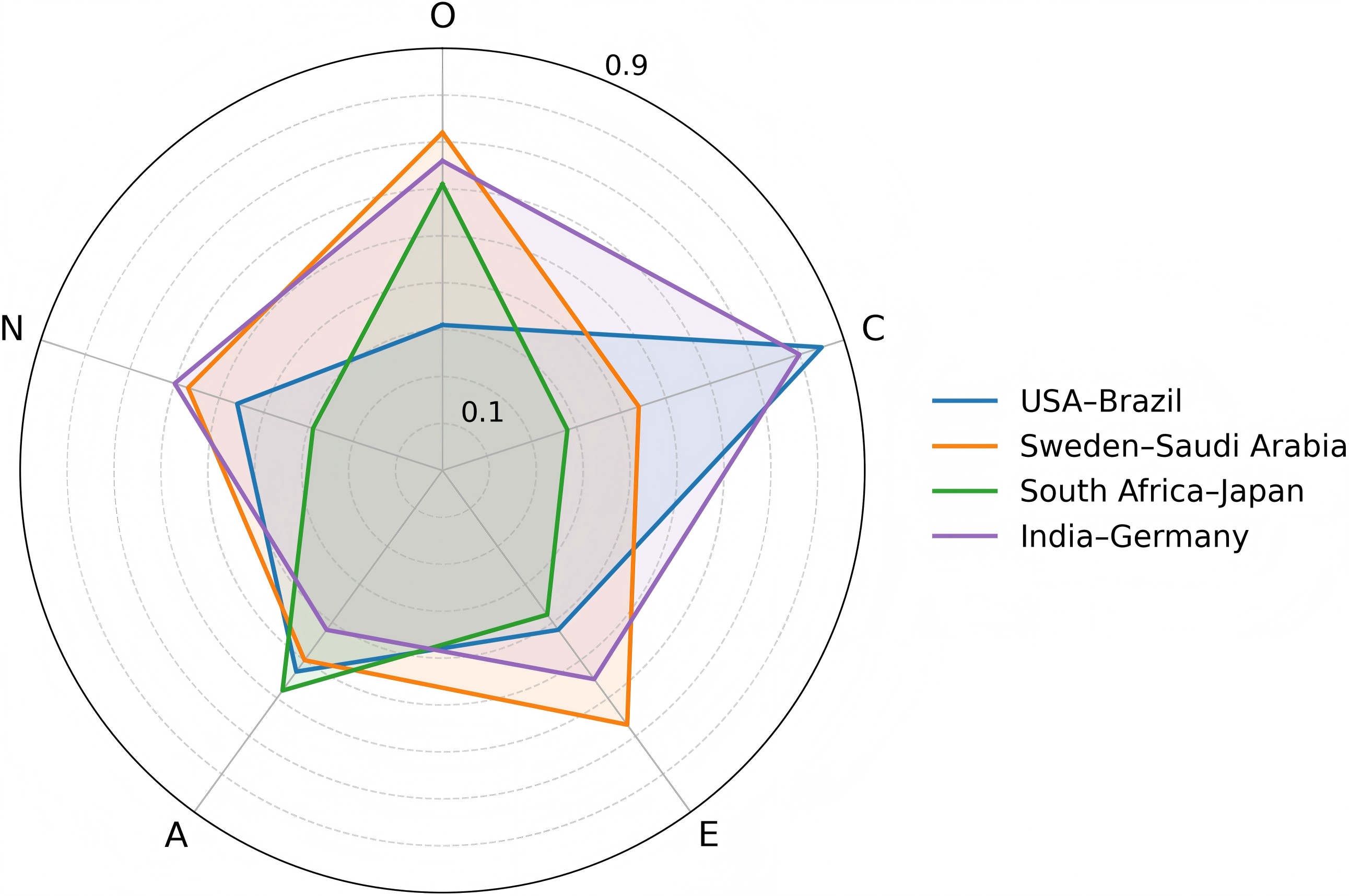}
    \caption{Pairwise Wasserstein distances for Big Five personality traits across four country pairs, with larger values indicating greater cross-cultural differences in trait distributions.}
    \label{fig:radar}
\end{figure}

Recent research has advanced the cultural awareness of large language models (LLMs) through the use of novel data resources and targeted training strategies. Culture-Gen \cite{li2024culture} proposes a prompting framework that systematically elicits cultural perceptions across countries and topics, enabling large-scale analysis of implicit cultural knowledge and bias without external knowledge bases or manual annotation. By contrast, CultureBank \cite{shi2024culturebank} constructs an open, structured knowledge base from culture-related narratives in online communities, using a pipeline that integrates automated filtering, LLM-assisted extraction, clustering, and human validation to capture fine-grained, community-driven cultural practices. CultureLLM \cite{li2024culturellm} further improves scalability by leveraging representative questions from the World Value Survey \cite{haerpfer2020world} and semantic data augmentation to efficiently expand training data for low-cost fine-tuning of both culture-specific and unified models.

Most existing approaches, however, tend to neglect the role of personality traits in shaping cross-cultural interactions \cite{pawar2025survey}. At the behavioural level, culture and personality are closely intertwined, with cultural context influencing both the expression and distribution of personality traits across groups. Cross-cultural studies \cite{mccrae2005personality,schmitt2007geographic} have shown that average levels of Big Five\footnote{The Big Five, also known as the Five-Factor Model, is the most widely accepted framework in personality psychology, describing five fundamental dimensions of human personality: Openness, Conscientiousness, Extraversion, Agreeableness, and Neuroticism (OCEAN).}~\cite{costa2008revised} traits vary substantially across countries and cultural contexts, a pattern further illustrated by the IPIP-NEO data \cite{johnson2014measuring} in Figure~\ref{fig:radar}. Personality is also systematically reflected in linguistic style \cite{yarkoni2010personality,shen2025ll4g,chen2025hippd}. For example, high openness is associated with greater curiosity and the use of abstract or imaginative vocabulary, as reflected in words such as ``poet'' or ``universe'', whereas high conscientiousness favours structured, goal-oriented expressions and avoids overly casual language. For conversational agents, incorporating personality traits therefore provides an effective means of modelling cultural and social context.

In this paper, we propose \textbf{HiCAP} (\textbf{Hi}erarchical \textbf{C}ultural \textbf{A}daptation via \textbf{P}ersonality), which explicitly integrates personality traits into culturally adaptive dialogue generation. The core of HiCAP lies in a personality-conditioned control mechanism that first establishes a culture aligned baseline for tone and politeness, and then refines linguistic style and content selection according to individual personality traits, thereby enabling hierarchical modulation of response style. Specifically, during training, supervised fine-tuning \cite{dong2023abilities} on dialogue data annotated with cultural and personality attributes allows the model to learn mappings between persona representations and stylistic adjustments through a gating mechanism. In addition, HiCAP employs slot-filling techniques \cite{mirza2024illuminer} to dynamically construct instruction templates, which are adaptively adjusted based on users' cultural background and personality traits. When explicit user persona information is unavailable, the framework infers the most likely cultural background and personality traits from the input, ensuring personalised instructions across different personality-culture combinations. Inspired by CulturePark \cite{li2024culturepark}, HiCAP further adopts multi-agent interaction to generate cross-cultural dialogue demonstrations, providing representative exemplars that illustrate the intended communicative style.

Our main contributions are as follows:

\begin{itemize}
    \item We propose HiCAP, a framework that integrates personality traits into culturally adaptive dialogue generation through hierarchical style modulation, allowing cultural context to guide overall response style while personality traits refine individual expression.

    \item We develop persona-aware learning and instruction mechanisms that combine personality-conditioned gating, dynamic instruction construction, persona inference, and multi-agent demonstrations to support dialogue generation aligned with cultural context and personality traits.

    \item Experiments on CulturalPersonas and CPED demonstrate that HiCAP outperforms state-of-the-art methods across various dialogue generation metrics. Further analyses validate its cultural adaptability, personality modelling, and linguistic quality.
\end{itemize}

\section{Related Work}

\noindent \textbf{Training-based Methods:} This category comprises three main paradigms that include pre-training from scratch on culturally rich corpora, continued pre-training on culture-specific data, and supervised fine-tuning with culturally annotated datasets. Models such as HyperClovaX \cite{yoo2024hyperclova} and PersianLLaMA \cite{abbasi2023persianllama} follow the first paradigm, achieving strong native-level cultural competence through large-scale region-specific training. To improve scalability, continued pre-training has been widely adopted to adapt general-purpose models to cultural domains, as exemplified by TaiwanLLM \cite{lin2023taiwan}, Komodo \cite{owen2024komodo}, SeaLLM \cite{nguyen2023seallms}, and AceGPT \cite{huang2023acegpt}. In parallel, supervised fine-tuning leverages labelled data capturing cultural, demographic, or social nuances, including Chinese culture-specific corpora \cite{bai2024coig}, the Indonesian Cendol dataset \cite{cahyawijaya2024cendol}, and value survey-based resources for hate speech detection \cite{zhou2023cultural}. However, These training-based approaches generally incur substantial computational cost \cite{shen2025less}.

\noindent \textbf{Inference-time Methods:} These methods adjust the generation process through prompt engineering that explicitly specifies cultural background, social roles, or value systems, without modifying model parameters. Representative strategies include culture-specific tokens or templates \cite{rao2024normad} and the explicit injection of cultural identity into prompts \cite{wang2023cdeval}. Chain-of-thought and value-oriented prompting further encourage reasoning within diverse moral or cultural frameworks \cite{johnson2022ghost, masoud2023cultural}. Recent studies have also shown that language-specific prompt designs can mitigate bias towards dominant cultures, particularly for underrepresented groups \cite{naous2023having}. In addition, prompt-based cultural alignment tests and context-aware evaluation prompts have been used to reveal implicit cultural preferences and assess cultural adaptability across tasks \cite{cao2023assessing, fung2024massively}. However, their capacity to support adaptation remains constrained when the underlying model lacks sufficient culture-specific knowledge.

HiCAP differs from both categories by jointly modelling culture and personality across training and inference. Persona-aware supervised fine-tuning embeds cultural and personality signals into the model's internal representations, while dynamic instructions and multi-agent demonstrations provide explicit, context-sensitive guidance during generation. A hierarchical personality-conditioned control mechanism further modulates response style by establishing a cultural baseline and refining it according to individual personality traits.

\section{HiCAP Framework}
\label{sec:method}

\begin{figure*}[tbp]
    \centering
    \includegraphics[width=\textwidth]{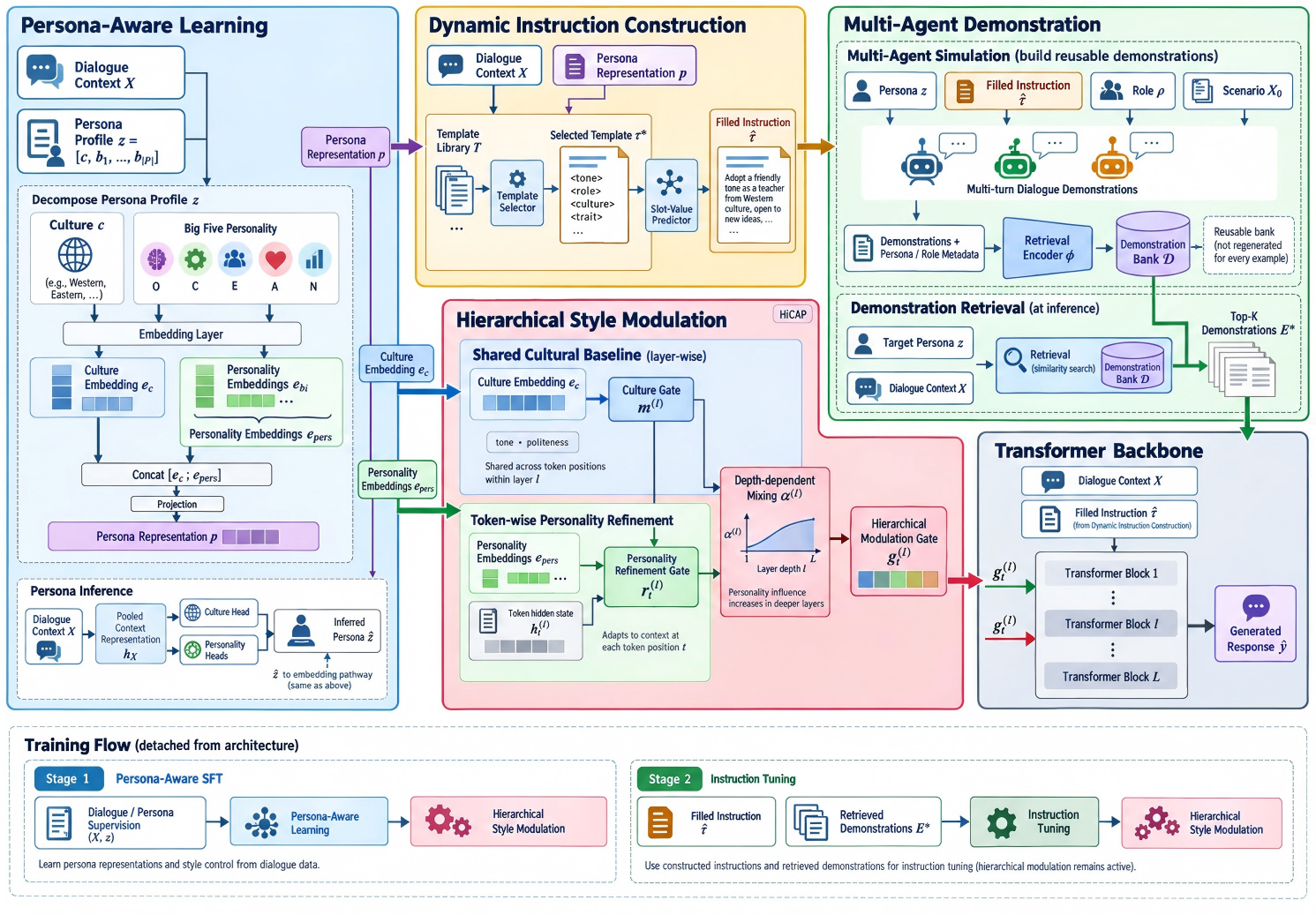}
    \caption{Overview of HiCAP, combining persona-aware learning, dynamic instruction construction, multi-agent demonstrations, and hierarchical style modulation to generate culturally and personality-aligned dialogue.}
    \label{fig:architecture}
\end{figure*}

As shown in Figure~\ref{fig:architecture}, HiCAP integrates personality traits into culturally adaptive dialogue generation through hierarchical personality-conditioned control. It combines persona-aware supervised fine-tuning (SFT), dynamic instruction construction, and multi-agent demonstrations. A hierarchical control mechanism uses cultural information to establish a shared stylistic baseline and personality information to refine expression at the token level.

\subsection{Persona-Aware Learning}
\label{sec:persona_learning}

Persona-aware learning represents cultural background and personality traits as conditioning signals during SFT, allowing persona information to influence the model's internal representations.

We represent each persona using a structured profile that combines a cultural category with binary indicators for the Big Five personality dimensions. Let $c$ denote the cultural category and $b_i \in \{0,1\}$ denote the state of the $i$-th personality trait, where $b_i=1$ and $b_i=0$ correspond to high and low levels, respectively. The persona profile is $\mathbf{z}=[c,b_1,b_2,\ldots,b_{|\mathcal{P}|}]$, where $\mathcal{P}$ denotes the set of personality dimensions. Each discrete attribute is mapped to a trainable embedding, yielding a culture embedding $\mathbf{e}_c$ and personality embeddings $\mathbf{e}_{b_i}$. These embeddings are concatenated and projected into the hidden space of the backbone as $\mathbf{p}=\mathbf{W}_{\mathrm{proj}}\operatorname{Concat}(\mathbf{e}_c,\mathbf{e}_{b_1},\ldots,\mathbf{e}_{b_{|\mathcal{P}|}})$, where $\mathbf{W}_{\mathrm{proj}}$ is a learnable projection matrix and $\mathbf{p}\in\mathbb{R}^{d_h}$ is the resulting persona representation with hidden dimension $d_h$. The projected representation is used for dynamic instruction construction, while the culture and personality embeddings are retained for hierarchical modulation.

During SFT, let $X$ denote the dialogue context and $Y=(y_1,\ldots,y_{|Y|})$ the target response. We optimise the standard next-token prediction loss $\mathcal{L}_{\mathrm{SFT}}=-\sum_{t=1}^{|Y|}\log P_{\Theta}(y_t\mid y_{<t},X,\mathbf{z})$, where $\Theta$ denotes the trainable model parameters.

To encourage the hidden representations to retain persona information, we additionally predict the cultural category and personality traits from a pooled context representation $\mathbf{h}_X$. The culture head produces $\hat{\mathbf{y}}_c=\operatorname{softmax}(\mathbf{W}_{\mathrm{cls},c}\mathbf{h}_X+\boldsymbol{\beta}_{\mathrm{cls},c})$, where $\mathbf{W}_{\mathrm{cls},c}$ and $\boldsymbol{\beta}_{\mathrm{cls},c}$ are its trainable weight matrix and bias vector. For the $i$-th personality dimension, the corresponding prediction head produces $\hat{q}_i=\sigma(\mathbf{w}_{\mathrm{cls},i}^{\top}\mathbf{h}_X+\beta_{\mathrm{cls},i})$, where $\mathbf{w}_{\mathrm{cls},i}$ and $\beta_{\mathrm{cls},i}$ are trainable parameters and $\sigma(\cdot)$ denotes the sigmoid function. We denote the resulting auxiliary persona classification loss by $\mathcal{L}_{\mathrm{cls}}=\ell_{\mathrm{cls}}(\mathbf{z},\mathbf{h}_X)$, where $\ell_{\mathrm{cls}}$ jointly measures errors in cultural-category and personality-trait prediction. This auxiliary supervision encourages the internal representation to remain aligned with the intended persona during training.

\subsection{Dynamic Instruction Construction}
\label{sec:dynamic_instruction}

Dynamic instruction construction converts the structured persona representation into explicit natural-language guidance for response generation. Let $\mathcal{T}$ denote a curated set of instruction template schemas with parameterised slots. A template $\tau\in\mathcal{T}$ is a sequence of ordinary tokens interleaved with semantic placeholders, such as $\langle\text{tone}\rangle$ or $\langle\text{role}\rangle$. Formally, $\tau=(t_1,\ldots,\langle\text{slot}_k\rangle,\ldots,t_{n_\tau})$, where $n_\tau$ is the number of elements in the template and $k$ indexes a slot. Each slot specifies a controllable attribute related to culture, personality, or interaction context. For the $k$-th slot, a candidate set $\mathcal{V}_k$ is constructed from high-frequency expressions in labelled dialogue corpora and manually curated variants to maintain stylistic coverage and linguistic naturalness.

Table~\ref{tab:template} (Appendix~\ref{app:template}) summarises the slot categories, slot types, and representative values used in the dynamic templates.

For a persona representation $\mathbf{p}$ and dialogue context $X$, each template receives a compatibility score $s_{\mathrm{temp}}(\tau\mid\mathbf{p},X)$ from a learned scoring function. The highest-scoring template is selected as $\tau^\ast=\arg\max_{\tau\in\mathcal{T}}s_{\mathrm{temp}}(\tau\mid\mathbf{p},X)$. Each slot in $\tau^\ast$ is then instantiated with a context-sensitive value $v_k\in\mathcal{V}_k$ selected by a learned slot-value predictor. The resulting instruction $\hat{\tau}=\operatorname{FillSlots}(\tau^\ast,\{v_k\}_k)$ expresses the relevant cultural, personality, and interaction guidance in natural language and is used as explicit conditioning during instruction tuning and generation.

When explicit persona annotations are unavailable, HiCAP infers the persona profile from the dialogue context. Using the prediction heads defined in Section~\ref{sec:persona_learning}, the cultural category is estimated as $\hat{c}=\arg\max_j\hat{y}_{c,j}$, where $\hat{y}_{c,j}$ is the predicted probability of cultural category $j$. Each personality indicator is obtained as $\hat{b}_i=\mathbb{I}[\hat{q}_i\geq\gamma]$, where $\gamma$ is a fixed decision threshold and $\mathbb{I}[\cdot]$ is the indicator function. The inferred profile $\hat{\mathbf{z}}=[\hat{c},\hat{b}_1,\ldots,\hat{b}_{|\mathcal{P}|}]$ is mapped to the persona representation $\mathbf{p}$ using the same embedding and projection functions as an annotated profile.

\subsection{Multi-Agent Demonstration}
\label{sec:multi_agent}

Inspired by CulturePark~\cite{li2024culturepark}, we use automated multi-agent simulation to generate culturally grounded and personality-consistent dialogue demonstrations. Each simulated agent is assigned a structured persona profile $\mathbf{z}$ and its corresponding dynamic instruction $\hat{\tau}$, allowing cultural background and personality traits to shape the simulated interaction.

Let $\rho\in\mathcal{V}_{\mathrm{role}}$ denote an interaction role, such as \texttt{advisor} or \texttt{friend}, where $\mathcal{V}_{\mathrm{role}}$ is the set of available roles. Given a scenario seed $X_0$, which provides a brief description of the interaction setting, the relevant role is instantiated in the corresponding instruction slot. The initial prompt for an agent is then constructed as $\Pi(\mathbf{z},\rho,X_0)=[\,\hat{\tau}\oplus C_{\mathrm{sys}}\oplus X_0\,]$, where $C_{\mathrm{sys}}$ denotes the system context and $\oplus$ denotes sequence concatenation.

During simulation, agents take turns generating utterances under their respective persona configurations, producing multi-turn dialogue trajectories across different cultural backgrounds, personality traits, roles, and scenarios. Persona conditioning remains active during generation, allowing these attributes to influence lexical choice and stylistic realisation throughout the interaction.

The generated demonstrations are stored with metadata describing their persona attributes, roles, and interaction contexts. For each demonstration $u$, let $\mathbf{a}_u=[c,b_1,\ldots,b_{|\mathcal{P}|},\rho]$ denote its structured metadata. Let $\phi(\cdot)$ denote a retrieval encoder that maps metadata and text into a shared semantic space. The demonstration is encoded as $\mathbf{e}_u=\phi(\mathbf{a}_u\oplus u)$.

Given a target persona--context pair $(\mathbf{z}^{\ast},X^{\ast})$, the query is encoded as $\mathbf{e}_q=\phi(\mathbf{z}^{\ast}\oplus X^{\ast})$ in the same semantic space. For each demonstration $u$ in the demonstration bank $\mathcal{D}$, the retrieval score is $s_{\mathrm{ret}}(u\mid\mathbf{z}^{\ast},X^{\ast})=\operatorname{sim}(\mathbf{e}_u,\mathbf{e}_q)$, where $\operatorname{sim}(\cdot,\cdot)$ denotes similarity in the embedding space. The $K$ highest-scoring demonstrations, where $K$ is the number of retrieved exemplars, form the set $\mathcal{E}^{\ast}$.

After persona-aware SFT, HiCAP performs instruction tuning with the filled instruction and retrieved demonstrations as additional conditioning. For a training pair $(X,Y,\mathbf{z})$, instruction tuning optimises the standard next-token loss $\mathcal{L}_{\mathrm{IT}}=-\sum_{t=1}^{|Y|}\log P_{\Theta}(y_t\mid y_{<t},X,\mathbf{z},\hat{\tau},\mathcal{E}^{\ast})$. At inference time, the same retrieval procedure supplies $\mathcal{E}^{\ast}$ as culturally and personality-aligned few-shot exemplars for response generation.

\subsection{Hierarchical Style Modulation}
\label{sec:hierarchical_modulation}

Hierarchical style modulation controls how cultural and personality information influence generation. Cultural information establishes a shared stylistic baseline, including properties such as tone and politeness, while personality traits refine expression according to the current token context.

Let $L$ denote the number of transformer blocks and $T$ the number of token positions. We retain the culture embedding $\mathbf{e}_c$ and combine the personality embeddings as $\mathbf{e}_{\mathrm{pers}}=\operatorname{Concat}(\mathbf{e}_{b_1},\ldots,\mathbf{e}_{b_{|\mathcal{P}|}})$. At transformer block $l\in\{1,\ldots,L\}$, the culture gate $\mathbf{m}^{(l)}=\sigma(F^{(l)}_{\mathrm{cult}}(\mathbf{e}_c))$ defines a baseline shared across token positions. For token position $t\in\{1,\ldots,T\}$, the personality refinement gate $\mathbf{r}^{(l)}_t=\sigma(F^{(l)}_{\mathrm{pers}}([\mathbf{h}^{(l)}_t\|\mathbf{e}_{\mathrm{pers}}]))$ adapts the modulation to the evolving hidden state $\mathbf{h}^{(l)}_t$, where $F^{(l)}_{\mathrm{cult}}$ and $F^{(l)}_{\mathrm{pers}}$ are trainable gate networks and $\|$ denotes vector concatenation.

We use a learned depth-dependent mixing coefficient $\alpha^{(l)}\in(0,1)$ that increases with layer depth, giving personality refinement progressively greater influence in deeper transformer blocks. The final token-wise modulation gate is

\begin{equation}
\mathbf{g}^{(l)}_t
=
\bigl(1-\alpha^{(l)}\bigr)\mathbf{m}^{(l)}
+
\alpha^{(l)}\mathbf{r}^{(l)}_t.
\label{eq:hier_gate}
\end{equation}

Using the gate in Equation~\ref{eq:hier_gate}, the hidden state is updated as

\begin{equation}
\mathbf{h}^{(l+1)}_t
=
\mathbf{g}^{(l)}_t
\odot
f^{(l)}_{\Theta}\!\left(\operatorname{LN}(\mathbf{h}^{(l)}_t)\right)
+
\bigl(\mathbf{1}-\mathbf{g}^{(l)}_t\bigr)
\odot
\mathbf{h}^{(l)}_t,
\label{eq:gated_update}
\end{equation}

where $f^{(l)}_{\Theta}(\cdot)$ denotes the transformation performed by block $l$, $\operatorname{LN}(\cdot)$ denotes layer normalisation, $\odot$ denotes element-wise multiplication, and $\mathbf{1}$ is an all-ones vector with the same dimensionality as the gate. The residual term preserves the contextual representation, while persona dependence enters through the gate controlling the transformed and residual states.

To keep personality refinement anchored to the cultural baseline, let $\bar{\mathbf{g}}^{(l)}=\frac{1}{T}\sum_{t=1}^{T}\mathbf{g}^{(l)}_t$ denote the gate averaged across token positions at layer $l$. We define the hierarchical regulariser as

\begin{equation}
\mathcal{L}_{\mathrm{hier}}
=
\sum_{l=1}^{L}
\left\|
\bar{\mathbf{g}}^{(l)}-\mathbf{m}^{(l)}
\right\|_2^2.
\label{eq:hier_loss}
\end{equation}

Equation~\ref{eq:hier_loss} penalises the average deviation of the token-wise modulation gate from the cultural baseline while retaining token-specific variation. Let $\mathcal{L}_{\mathrm{gen}}$ denote $\mathcal{L}_{\mathrm{SFT}}$ during persona-aware SFT and $\mathcal{L}_{\mathrm{IT}}$ during instruction tuning. Combining the corresponding next-token loss, auxiliary persona supervision, and hierarchical regularisation, the training objective for each stage is

\begin{equation}
\mathcal{L}_{\mathrm{total}}
=
\mathcal{L}_{\mathrm{gen}}
+
\lambda_{\mathrm{cls}}\mathcal{L}_{\mathrm{cls}}
+
\lambda_{\mathrm{hier}}\mathcal{L}_{\mathrm{hier}},
\label{eq:total_loss}
\end{equation}

where $\lambda_{\mathrm{cls}}$ and $\lambda_{\mathrm{hier}}$ control the contributions of persona supervision and hierarchical regularisation, respectively.

\begin{algorithm}[tbp]
\caption{HiCAP Training and Inference}
\label{alg:pseudocode}
\begin{algorithmic}[1]
\Require Context $X$, persona $\mathbf{z}$, optional target $Y$, demonstration bank $\mathcal{D}$, parameters $\Theta$, learning rate $\eta$
\If{$Y$ is available}
    \State Compute $\mathcal{L}_{\mathrm{total}}$ on $(X,Y,\mathbf{z})$ with layer-wise hierarchical modulation and $\mathcal{L}_{\mathrm{gen}}=\mathcal{L}_{\mathrm{SFT}}$
    \State Update $\Theta$ using $\mathcal{L}_{\mathrm{total}}$ with learning rate $\eta$
    \State $(\mathbf{p},\mathbf{e}_c,\mathbf{e}_{\mathrm{pers}})\leftarrow\textsc{Embed}(\mathbf{z})$
    \State $\tau^{\ast}\leftarrow\textsc{Template}(\mathbf{p},X)$; $\hat{\tau}\leftarrow\textsc{Fill}(\tau^{\ast},\mathbf{p},X)$
    \State $\mathcal{E}^{\ast}\leftarrow\textsc{Retrieve}(\mathcal{D},\mathbf{z},X)$
    \State Compute $\mathcal{L}_{\mathrm{total}}$ on $(X,\hat{\tau},\mathcal{E}^{\ast},Y,\mathbf{z})$ with layer-wise hierarchical modulation and $\mathcal{L}_{\mathrm{gen}}=\mathcal{L}_{\mathrm{IT}}$
    \State Update $\Theta$ using $\mathcal{L}_{\mathrm{total}}$ with learning rate $\eta$
\Else
    \State Infer $\mathbf{z}\leftarrow\textsc{Infer}(X)$ if unavailable
    \State $(\mathbf{p},\mathbf{e}_c,\mathbf{e}_{\mathrm{pers}})\leftarrow\textsc{Embed}(\mathbf{z})$
    \State $\tau^{\ast}\leftarrow\textsc{Template}(\mathbf{p},X)$; $\hat{\tau}\leftarrow\textsc{Fill}(\tau^{\ast},\mathbf{p},X)$
    \State $\mathcal{E}^{\ast}\leftarrow\textsc{Retrieve}(\mathcal{D},\mathbf{z},X)$
    \State $\hat{Y}\leftarrow\textsc{Decode}(X,\hat{\tau},\mathcal{E}^{\ast})$ with layer-wise hierarchical modulation
    \State \Return $\hat{Y}$
\EndIf
\end{algorithmic}
\end{algorithm}

Algorithm~\ref{alg:pseudocode} summarises how the components described above are integrated in the training and inference processes of HiCAP.

\section{Experiments and Results}

\subsection{Experimental Setup}
\label{sec:experimental_setup}

\noindent \textbf{Datasets:} 
We adopt CulturalPersonas \cite{dey2025can} to evaluate personality expression across different cultural contexts and CPED \cite{chen2022cped} to examine personality expression and cultural adaptation across different social groups within a single society. Together, they provide complementary cross-cultural and intra-cultural evaluation settings. CulturalPersonas is a large-scale, human-validated benchmark designed to assess how LLMs express personality traits across authentic cultural contexts. It consists of 3,000 scenario-based question--answer pairs from six countries (United States, Brazil, India, Japan, Saudi Arabia, and South Africa). Each instance is associated with one of the Big Five personality dimensions and supports both Likert-style multiple-choice and open-ended evaluation. CPED is a large-scale Chinese dialogue dataset containing over 133,000 utterances across more than 12,000 multi-turn dialogues extracted from 40 contemporary Chinese TV dramas. Utterances are annotated with sentiment, emotion, dialogue acts, scene categories, and static personal attributes, including gender, age, and Big Five traits.

\noindent \textbf{Implementation Details:}
We use the BPDD dataset \cite{liu2025bilingual} for supervised fine-tuning and Phi-4-14B \cite{abdin2024phi} as the backbone. The model is trained in bfloat16 precision with a maximum context length of 16K tokens on an NVIDIA H200 cluster. Both training stages use the AdamW optimiser with a cosine learning-rate schedule. Newly introduced modules use a learning rate of $2 \times 10^{-5}$, while the backbone uses $5 \times 10^{-6}$. The dynamic instruction library contains 36 template schemas with 4--9 parameterised slots each. Culture and personality embeddings have dimensions of 128 and 64, respectively. The culture and personality gate networks use two-layer MLPs with hidden size 2048, while the slot-value predictor uses an MLP with hidden size 512; all newly introduced MLPs use GELU activation. The demonstration index contains approximately 1.2 million persona-aligned utterances, with synthetic demonstrations capped at 30\% of the conditioning tokens during instruction tuning to mitigate overfitting. Decoding uses nucleus sampling with $p=0.95$ and temperature $0.7$.

\begin{table*}[htbp]
\centering
\scalebox{0.75}{
\begin{tabular}{llcccccc}
\hline
\textbf{Dataset} & \textbf{Model} & \textbf{BLEU} $\uparrow$ & \textbf{BERTScore} $\uparrow$ & \textbf{Perplexity} $\downarrow$ & \textbf{TTR} $\uparrow$ & \textbf{Distinct-1} $\uparrow$ & \textbf{Distinct-2} $\uparrow$ \\
\hline
\multirow{9}{*}{\textbf{CPED}} 
& Seq2Seq & 0.0077 & 0.5196  & 107.30 & 0.4342 & 0.0252 & 0.1846 \\
& Transformer & 0.1680 & 0.5519 & 62.82 & 0.5687 & 0.0264 & 0.2031 \\
& CDial-GPT & 0.1403 & 0.5719 & 17.70 & 0.6412 & 0.0602 & 0.3388 \\
& Qwen-2.5-72B-Instruct & 0.1710 & 0.6815 & 15.50 & 0.6910 & 0.0707 & 0.3587 \\
& Llama-3.3-70B-Instruct & 0.1621 & 0.6705 & 15.92 & 0.6820 & 0.0663 & 0.3451 \\
& Gemma-2-27B-IT & 0.1553 & 0.6642 & 16.88 & 0.6755 & 0.0646 & 0.3385 \\
& DeepSeek-V3.1 & 0.1748 & 0.6889 & 15.15 & 0.6928 & 0.0723 & 0.3640 \\
& GPT-4o & 0.1821 & 0.7035 & 14.88 & 0.6994 & 0.0752 & 0.3743 \\
& \textbf{HiCAP (Ours)} & \textbf{0.1884} & \textbf{0.7112} & \textbf{14.72} & \textbf{0.7028} & \textbf{0.0761} & \textbf{0.3812} \\
\hline
\multirow{7}{*}{\textbf{CulturalPersonas}} 
& Qwen-2.5-72B-Instruct    & 0.1639 & 0.6764 & 16.38 & 0.7162 & 0.0773 & 0.3776 \\
& Llama-3.3-70B-Instruct  & 0.1612 & 0.6727 & 16.22 & 0.7141 & 0.0763 & 0.3731 \\
& Gemma-2-27B-IT          & 0.1576 & 0.6683 & 16.66 & 0.7083 & 0.0742 & 0.3671 \\
& Kimi-K2-Instruct        & 0.1648 & 0.6849 & 16.03 & 0.7256 & 0.0823 & 0.3909 \\
& DeepSeek-V3.1           & 0.1677 & 0.6814 & 15.88 & 0.7223 & 0.0804 & 0.3862 \\
& GPT-4o                  & 0.1714 & 0.6936 & \textbf{15.62} & 0.7281 & 0.0842 & \textbf{0.4023} \\
& \textbf{HiCAP (Ours)}          & \textbf{0.1762} & \textbf{0.7001} & 15.78 & \textbf{0.7309} & \textbf{0.0853} & 0.3919 \\
\hline
\end{tabular}
}
\caption{Dialogue generation performance of different models on CPED and CulturalPersonas. $\uparrow$ indicates higher is better; $\downarrow$ indicates lower is better.}
\label{tab:merged_eval}
\end{table*}

\noindent \textbf{Evaluation Metrics:}  
Following the evaluation settings of CulturalPersonas and CPED, we adopt several complementary evaluation metrics, including BLEU for quantifying surface-level n-gram overlap, BERTScore for measuring semantic similarity between model outputs and reference texts using contextual embeddings, Perplexity for assessing generation fluency and predictive confidence, and the type--token ratio (TTR) together with Distinct-$n$ for capturing vocabulary richness and variability, where TTR focuses on the unique word ratio and Distinct-$n$ extends this to n-gram diversity. These metrics are jointly well suited for cross-cultural evaluation, as cultural differences are often reflected in variations in lexical choice, semantic framing, and stylistic diversity.

\noindent \textbf{Comparative Models:} 
For CPED, we include Seq2Seq \cite{sutskever2014sequence}, Transformer \cite{vaswani2017attention}, and CDial-GPT \cite{wang2020large} models fine-tuned on the dataset. We further compare HiCAP with several advanced LLMs, including Qwen-2.5 \cite{team2024qwen2}, Llama-3.3 \cite{dubey2024llama}, Gemma-2 \cite{team2024gemma}, DeepSeek-V3.1 \cite{liu2024deepseek}, and GPT-4o \cite{hurst2024gpt}. Kimi-K2 \cite{team2025kimi} is additionally included for CulturalPersonas.

\subsection{Overall Results}

Table~\ref{tab:merged_eval} reports dialogue generation performance on CPED and CulturalPersonas. On CPED, HiCAP achieves the best performance across all six reported metrics. On CulturalPersonas, HiCAP achieves the highest BLEU, BERTScore, TTR, and Distinct-1, while GPT-4o obtains lower Perplexity and higher Distinct-2. The consistency of the gains across two substantially different dialogue settings suggests that HiCAP remains effective under both Chinese multi-turn dialogue and cross-cultural scenario-based generation.

The metric-level results reveal several distinct patterns. Improvements in both BLEU and BERTScore indicate that HiCAP increases agreement with reference responses at both the lexical and semantic levels, with BERTScore gains showing that the improvement is not confined to surface-form overlap. On CPED, HiCAP also achieves the lowest Perplexity while simultaneously obtaining the highest TTR and Distinct scores, indicating that greater lexical diversity is achieved without an accompanying degradation in sequence likelihood. CulturalPersonas presents a slightly different pattern. HiCAP attains the highest TTR and Distinct-1, suggesting broader lexical usage, whereas GPT-4o achieves the highest Distinct-2 and slightly lower Perplexity. This difference suggests that HiCAP favours vocabulary-level diversity, while GPT-4o exhibits greater diversity in local bigram combinations on this benchmark.

The comparison across model families further highlights the relative strength of HiCAP. Conventional dialogue models, including Seq2Seq, Transformer, and CDial-GPT, remain substantially behind recent LLMs, particularly in semantic similarity and lexical diversity. Among the general-purpose LLMs, GPT-4o remains the strongest baseline on most metrics across the two benchmarks, with models such as DeepSeek-V3.1 following closely, while Kimi-K2 also demonstrates strong lexical diversity. Despite these competitive baselines, HiCAP achieves the strongest overall performance across both benchmarks. Notably, HiCAP uses a 14B backbone while competing with substantially larger models ranging from 27 billion to 1 trillion parameters, suggesting that the observed gains cannot be explained by model scale alone.

\begin{table}[tbp]
\centering
\resizebox{\columnwidth}{!}{
\begin{tabular}{lccc}
\toprule
\textbf{Component} & \textbf{D-1} $\uparrow$ & \textbf{D-2} $\uparrow$ & \textbf{PPL} $\downarrow$ \\
\midrule
Vanilla Phi-4 & 0.0697 & 0.3228 & 18.04 \\
Replace Phi-4 with Qwen-2.5-14B & 0.0850 & 0.3913 & \textbf{15.64} \\
Replace Phi-4 with Llama-3.1-8B & 0.0823 & 0.3754 & 16.29 \\
w/o Persona-Conditioned Gating & 0.0729 & 0.3397 & 17.04 \\
Replace Token-Wise with Scalar & 0.0761 & 0.3538 & 16.93 \\
w/o Personality Inference & 0.0784 & 0.3602 & 16.47 \\
w/o Dynamic Template Selector & 0.0738 & 0.3441 & 16.52 \\
w/o Demonstrations & 0.0792 & 0.3653 & 15.98 \\
w/o Instruction Tuning & 0.0733 & 0.3426 & 17.18 \\
w/o Supervised Fine-Tuning & 0.0764 & 0.3544 & 16.91 \\
\midrule
\textbf{HiCAP} & \textbf{0.0853} & \textbf{0.3919} & 15.78 \\
\bottomrule
\end{tabular}
}
\caption{Ablation results for HiCAP on CulturalPersonas. $\uparrow$ indicates higher is better; $\downarrow$ indicates lower is better.}
\label{tab:ablation}
\end{table}

\subsection{Ablation Study}

We conduct an ablation study of the key components of HiCAP, as shown in Table~\ref{tab:ablation}. We focus on three complementary generation metrics: Distinct-1 (D-1), Distinct-2 (D-2), and Perplexity (PPL), which jointly assess lexical diversity and fluency.

We first examine the impact of replacing the backbone language model. Using the vanilla Phi-4 model without HiCAP components yields the lowest diversity and the highest perplexity, indicating the contribution of the proposed components to generation quality. Replacing the Phi-4 backbone with Qwen-2.5-14B achieves comparable diversity with slightly lower perplexity, while Llama-3.1-8B produces intermediate results. This indicates that the framework generalises across different backbones, while the stronger performance of the larger models may also reflect known scaling effects \cite{kaplan2020scaling}.

The removal of persona-conditioned gating causes a clear drop in both diversity metrics and an increase in perplexity, highlighting its central role in integrating personality and cultural features into the backbone model. Without this mechanism, the stylistic control learned from these signals is largely lost. Replacing token-wise gating with a scalar alternative likewise reduces diversity and increases perplexity, although the degradation is less severe than when gating is entirely absent. This comparison confirms that fine-grained, token-level control is more effective than a single global coefficient for modulating the influence of personality and cultural information during generation.

Eliminating the personality inference branch results in moderate reductions in both diversity and fluency. Even when cultural information is preserved, the lack of personality predictions limits the model’s ability to vary stylistic and lexical choices, particularly when explicit personality labels are unavailable in the input. Disabling the dynamic template selector leads to a more pronounced drop in diversity, as the model is constrained to a single static template with fixed slot values, reducing its adaptability across contexts. By contrast, demonstrations exert a relatively limited impact on generation quality. Although they facilitate alignment with context-specific styles, the backbone retains strong generation capability in their absence.

The final ablations examine the two post-training stages of HiCAP. Removing instruction tuning leads to a clear decrease in both diversity metrics and an increase in perplexity, underscoring its contribution to context-sensitive generation while maintaining fluency. Persona-aware supervised fine-tuning also contributes substantially, although its removal results in a smaller degradation across all three metrics. This pattern suggests that both post-training stages are important, with instruction tuning contributing more strongly to the measured diversity and fluency of the final model.

\subsection{Additional Analyses}

\subsubsection{Implicit Personality Recognition}

To evaluate the personality inference capability of HiCAP when explicit persona annotations are unavailable, we assess its ability to recover the Big Five traits directly from dialogue context on the CPED dataset. In this setting, recognition depends on subtle linguistic cues, pragmatic tendencies, and context-dependent expressions because explicit trait descriptions are absent \cite{shen2025emoperso}.

\begin{table}[htbp]
\centering
\resizebox{\columnwidth}{!}{
\begin{tabular}{lcccccc|c}
\hline
\textbf{Methods} & \textbf{AGR} & \textbf{CON} & \textbf{EXT} & \textbf{NEU} & \textbf{OPN} & \textbf{Avg} & \textbf{F1} \\
\hline
SVM & 70.83 & 57.90 & 60.89 & 47.76 & 49.47 & 57.37 & 54.74 \\
LSTM & 70.01 & 60.56 & 64.73 & 53.75 & 54.93 & 60.80 & 59.81 \\
BERT & 75.89 & 63.48 & 68.21 & 52.67 & 55.42 & 63.13 & 61.47 \\
RoBERTa & 78.08 & 63.72 & 67.65 & 53.60 & 55.58 & 63.73 & 61.98 \\
TrigNet & 82.76 & 66.52 & 69.15 & 58.52 & 54.50 & 66.29 & 63.45 \\
PS-GCN & 82.87 & 67.75 & 72.48 & 58.50 & 56.60 & 67.64 & 64.80 \\
GPT-4o & 80.16 & 65.40 & 71.59 & 59.80 & 53.29 & 66.05 & 63.54 \\
PsyCoT & 80.18 & 65.83 & 69.43 & 53.35 & 56.15 & 64.99 & 62.77 \\
EERPD & 84.03 & 73.91 & 74.29 & 52.67 & 56.44 & 68.27 & 67.06 \\
\hline
\textbf{HiCAP} & \textbf{85.36} & \textbf{75.18} & \textbf{74.92} & \textbf{60.24} & \textbf{57.31} & \textbf{70.60} & \textbf{68.74} \\
\hline
\end{tabular}
}
\caption{Performance comparison (\%) for Big Five personality recognition on CPED.}
\label{tab:personality}
\end{table}

For this task, we compare against classic machine learning and neural models such as SVM \cite{tighe2016personality}, LSTM \cite{tandera2017personality}, BERT \cite{devlin2019bert}, and RoBERTa \cite{liu2019roberta}, together with methods developed specifically for Big Five personality detection, including TrigNet \cite{yang2021psycholinguistic}, PS-GCN \cite{liu2024ps}, PsyCoT \cite{yang2023psycot}, and EERPD \cite{li2024eerpd}. As shown in Table~\ref{tab:personality}, HiCAP achieves the highest average accuracy and Macro-F1 among all compared methods, outperforming EERPD by 2.33\% and 1.68\%, respectively. The gains across all five traits suggest that the personality inference mechanism can reliably recover personality information from implicit linguistic and contextual signals even when explicit trait descriptions are unavailable.

\subsubsection{Cultural Priming Effects}

Cultural priming explicitly provides cultural cues at inference time, allowing the model to condition generation on a specified cultural frame. Although HiCAP already infers and encodes cultural information internally, explicit cultural cues may further reduce uncertainty in culturally conditioned expression.

\begin{figure}[tb]
    \centering
    \includegraphics[width=\columnwidth]{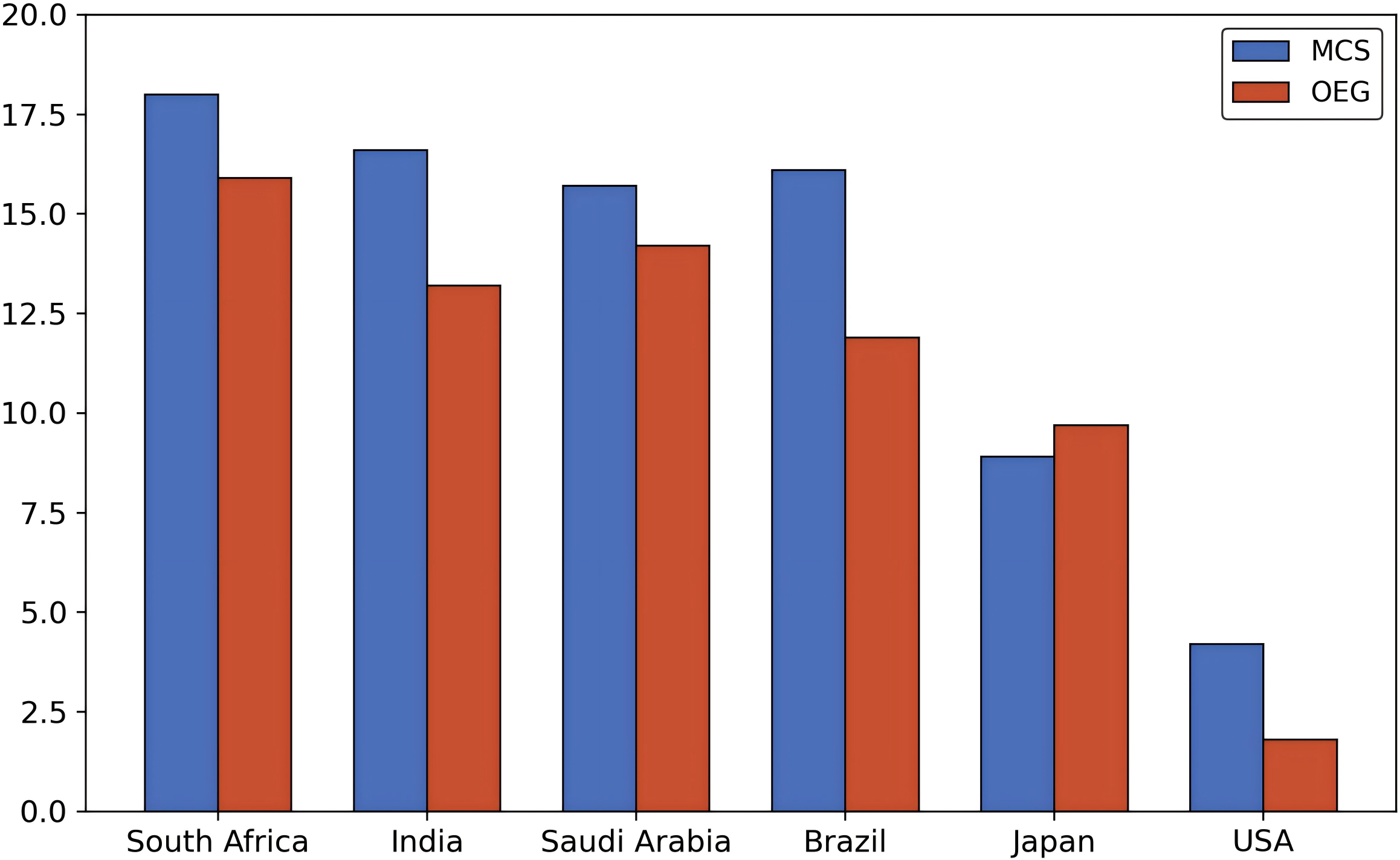}
    \caption{Cross-country comparison of HiCAP's improvement in cultural--personality alignment from cultural priming on CulturalPersonas.}
    \label{fig:priming}
\end{figure}

As shown in Figure~\ref{fig:priming}, cultural priming produces additional improvements under both the Likert-style multiple-choice (MCS) and open-ended generation (OEG) settings on CulturalPersonas, measured by relative reductions in Wasserstein distance. The largest marginal gains are observed in South Africa, India, and Saudi Arabia. This pattern may reflect greater divergence between these cultural contexts and dominant pretraining distributions, making explicit cultural grounding more beneficial. The gains in the United States are smaller, possibly because HiCAP's internal cultural inference is already closer to the target cultural priors without additional prompting. These results suggest that cultural priming provides greater benefit when the model's internal cultural representation is less closely aligned with the target context.

\subsubsection{Automatic Linguistic Evaluation}
\label{sec:linguistic_eval}

To evaluate the linguistic quality of HiCAP's generated dialogues, we use an LLM-as-a-judge framework \cite{li2024llms} on CulturalPersonas and CPED. The framework assesses fluency and coherence based on grammatical naturalness, semantic appropriateness, and contextual consistency within the dialogue. LLM-based evaluation has been widely used for assessing generated text and has shown strong agreement with human linguistic judgements \cite{zheng2023judging}.

Specifically, three external evaluators,
GPT-5 \cite{singh2025openai},
Claude-Sonnet-4.5\footnote{\href{https://assets.anthropic.com/m/12f214efcc2f457a/original/Claude-Sonnet-4-5-System-Card.pdf}{Claude-Sonnet-4.5 System Card}},
and Gemini-2.5-Pro \cite{comanici2025gemini},
are employed to assess HiCAP's outputs using a unified evaluation prompt, provided in Appendix~\ref{app:flu_prompt} and shown in Figure~\ref{fig:flu}.

\begin{table}[tb]
\centering
\scalebox{0.80}{
\begin{tabular}{lcc}
\toprule
\textbf{Evaluator} & \textbf{Fluency (\%)} & \textbf{Coherence (\%)} \\
\midrule
GPT-5                 & 98.8 & 98.5 \\
Claude-Sonnet-4.5     & 98.3 & 98.2 \\
Gemini-2.5-Pro        & 98.1 & 97.8 \\
\midrule
\textbf{Macro Average} & \textbf{98.4} & \textbf{98.2} \\
\bottomrule
\end{tabular}}
\caption{Evaluation of HiCAP's dialogue quality by three external LLM-based evaluators.}
\label{tab:flu_coh}
\end{table}

As shown in Table~\ref{tab:flu_coh}, HiCAP achieves average fluency and coherence scores of 98.4\% and 98.2\%, respectively. GPT-5 yields the highest scores, followed closely by Claude-Sonnet-4.5 and Gemini-2.5-Pro. The small differences across evaluators may reflect variations in their sensitivity to stylistic diversity and discourse continuity. Their closely aligned scores indicate consistent assessments of HiCAP's output quality. The slightly lower coherence scores may arise from conversational fillers and elliptical constructions that occur naturally in dialogue, while the high average scores indicate that the generated responses remain generally fluent and contextually coherent.

\subsection{Case Analysis}

Qualitative analyses of culturally adaptive response generation and safety risks are provided in Appendix~\ref{app:cultural_response} and Appendix~\ref{app:safety}, respectively.


\section{Conclusions}
We proposed HiCAP, a framework for culturally adaptive dialogue generation that uses personality traits to capture individual variation in cultural expression. HiCAP combines hierarchical personality-conditioned control with persona-aware learning, dynamic instruction construction, and multi-agent demonstrations. Experiments on CulturalPersonas and CPED demonstrate strong cultural adaptability, personality expressiveness, and overall generation quality, while ablation studies validate the contributions of the key components. Future work will explore finer-grained cultural and personality representations and extend HiCAP to multilingual and multimodal settings.

\clearpage

\section{Limitations}

HiCAP has several limitations. Cultural awareness in dialogue systems remains dependent on the availability and diversity of training data. Current corpora are heavily skewed towards high-resource and English-centric settings, with limited coverage of low-resource languages, regional dialects, and subcultural contexts. Extending beyond text to culturally representative images and videos could further strengthen contextual grounding and understanding of local norms, helping to mitigate cultural bias and improve global applicability.

More refined cultural modelling is also required. Many existing approaches rely on coarse labels such as country or language, which are insufficient to capture intra-cultural variation. Integrating psychological personality models with anthropological frameworks may support richer cultural representations. Such dimensions should be theoretically grounded and empirically meaningful. Combining fine-grained cultural factors with personality traits may enable more precise and context-sensitive adaptation without oversimplification.

Another limitation concerns the dynamic nature of culture. Current language models have limited capacity to continuously update their cultural knowledge as linguistic norms and social practices evolve. Continual learning or online adaptation could enable models to absorb emerging patterns while preserving accumulated knowledge. Human--AI co-adaptation based on feedback from specific communities may further support evolving cultural competence.

Finally, personality modelling remains challenging in terms of stability and reliability. Minor prompt variations can lead to substantial shifts in inferred personas, and certain personality configurations may increase the likelihood of unsafe or impolite outputs. More robust personality control mechanisms are needed to maintain behavioural coherence across contexts while balancing expressive flexibility with safety constraints.

\section{Ethical Considerations}

Personality and culture are deeply interconnected yet highly variable across individuals and societies. Treating them as fixed or tightly coupled attributes risks reinforcing stereotypes and obscuring within-culture diversity. Apparent differences in self-reported traits may reflect social norms or measurement bias rather than genuine psychological variation. Hard-coding ``typical personalities'' for specific cultures can therefore produce homogeneous responses that fail to capture individual differences. Culturally and personally adaptive systems should model these factors probabilistically and contextually, preserving flexibility across individuals and situations.

The integration of cultural and personality signals raises concerns about bias and privacy. Persona conditioning may amplify social stereotypes or reproduce biases present in training data, such as associating gender or nationality with particular traits. Datasets annotated with cultural and personality information may also contain sensitive personal data. Responsible development therefore requires safeguards for data collection, anonymisation, and use, together with transparent evaluation procedures to identify and mitigate bias.

Personalisation and diversity must also be balanced with fairness and user autonomy. Techniques such as on-device learning and federated aggregation can support user-specific adaptation while reducing the risk of data exposure. Users should retain explicit control over interaction styles and personality settings, ensuring that personalisation remains transparent and consensual. Evaluation frameworks should also emphasise multi-perspective alignment, allowing dialogue agents to represent diverse cultural viewpoints without enforcing a single normative stance.

\section*{Acknowledgments}

This work was supported by the Turing’s Defence and Security programme through a partnership with the UK government in accordance with the framework agreement between GCHQ \& The Alan Turing Institute.

\bibliography{custom}

\appendix

\section{Instruction Template Schema}
\label{app:template}

\begin{table}[htbp]
\centering
\resizebox{\columnwidth}{!}{
\begin{tabular}{c|l|l}
\hline
\textbf{Slot Category} & \textbf{Slot Type} & \textbf{Example Values} \\
\hline
\multirow{4}{*}{\centering Cultural background}
& \texttt{<culture>} & Japanese, British, Arabic \\
& \texttt{<culture\_adj>} & Japanese, British, Arabic (adjective form) \\
& \texttt{<cultural\_norm>} & values politeness, encourages directness \\
& \texttt{<language>} & Japanese, English, Spanish \\
\hline
\multirow{4}{*}{\centering Personality traits}
& \texttt{<trait\_high>} & high agreeableness, high conscientiousness \\
& \texttt{<trait\_low>} & low neuroticism, low openness \\
& \texttt{<trait\_behavior>} & calm and friendly, meticulous and careful \\
& \texttt{<trait\_emotion>} & reserved and modest, enthusiastic and expressive \\
\hline
\multirow{4}{*}{\centering Role and task context}
& \texttt{<role>} & assistant, advisor, friend \\
& \texttt{<task>} & answer questions, provide suggestions \\
& \texttt{<tone>} & formal, casual, humorous, warm \\
& \texttt{<formality>} & formal expressions, conversational style \\
\hline
\end{tabular}
}
\caption{Slot types and example values used in dynamic instruction templates.}
\label{tab:template}
\end{table}

\section{Prompt for Linguistic Evaluation}
\label{app:flu_prompt}

\begin{figure}[htbp]
    \centering
    \includegraphics[width=\columnwidth]{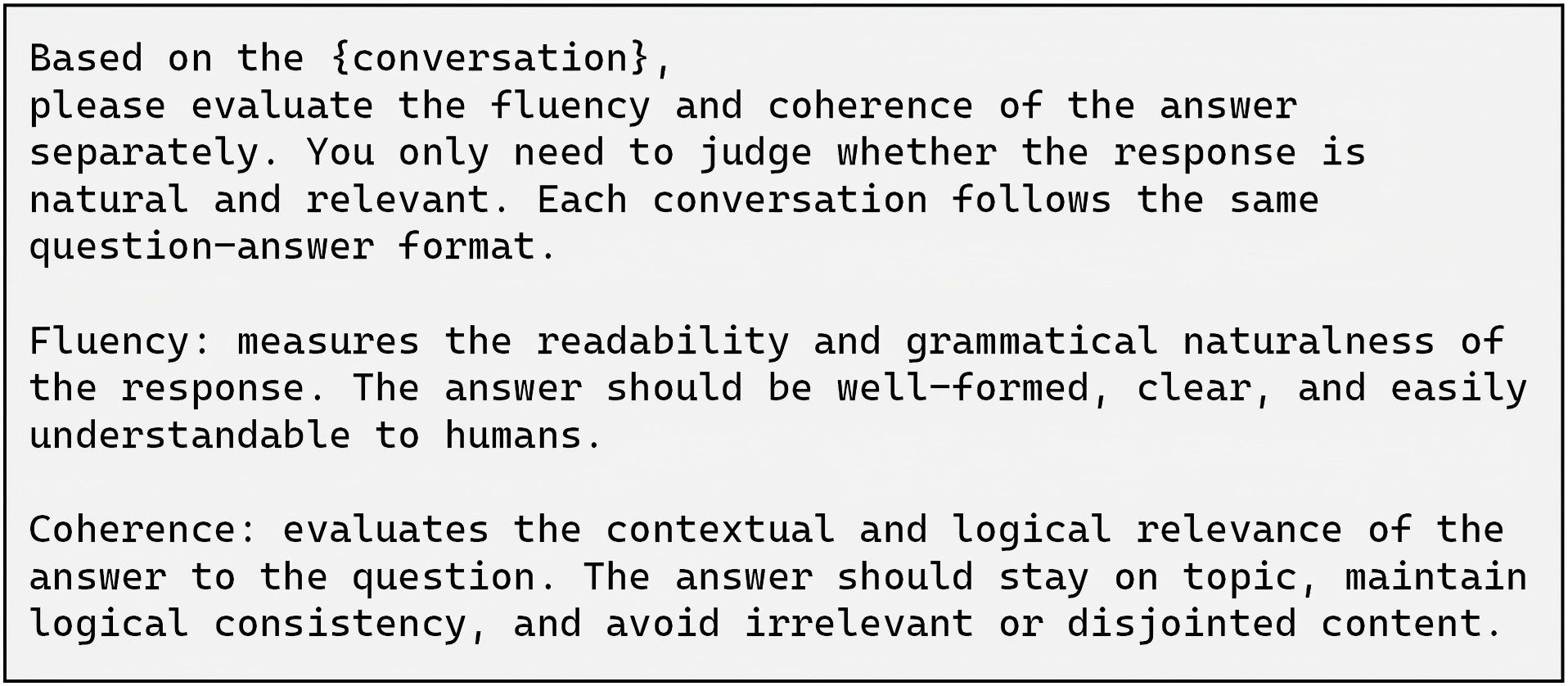}
    \caption{Unified prompt used to evaluate the fluency and coherence of HiCAP's generated dialogues.}
    \label{fig:flu}
\end{figure}

To ensure consistency and fairness across different LLM judges, we use a unified prompt to assess the fluency and coherence of HiCAP-generated dialogues. Figure~\ref{fig:flu} presents the complete prompt used in the evaluation.

\begin{figure*}[t]
    \centering
    \includegraphics[scale=0.195]{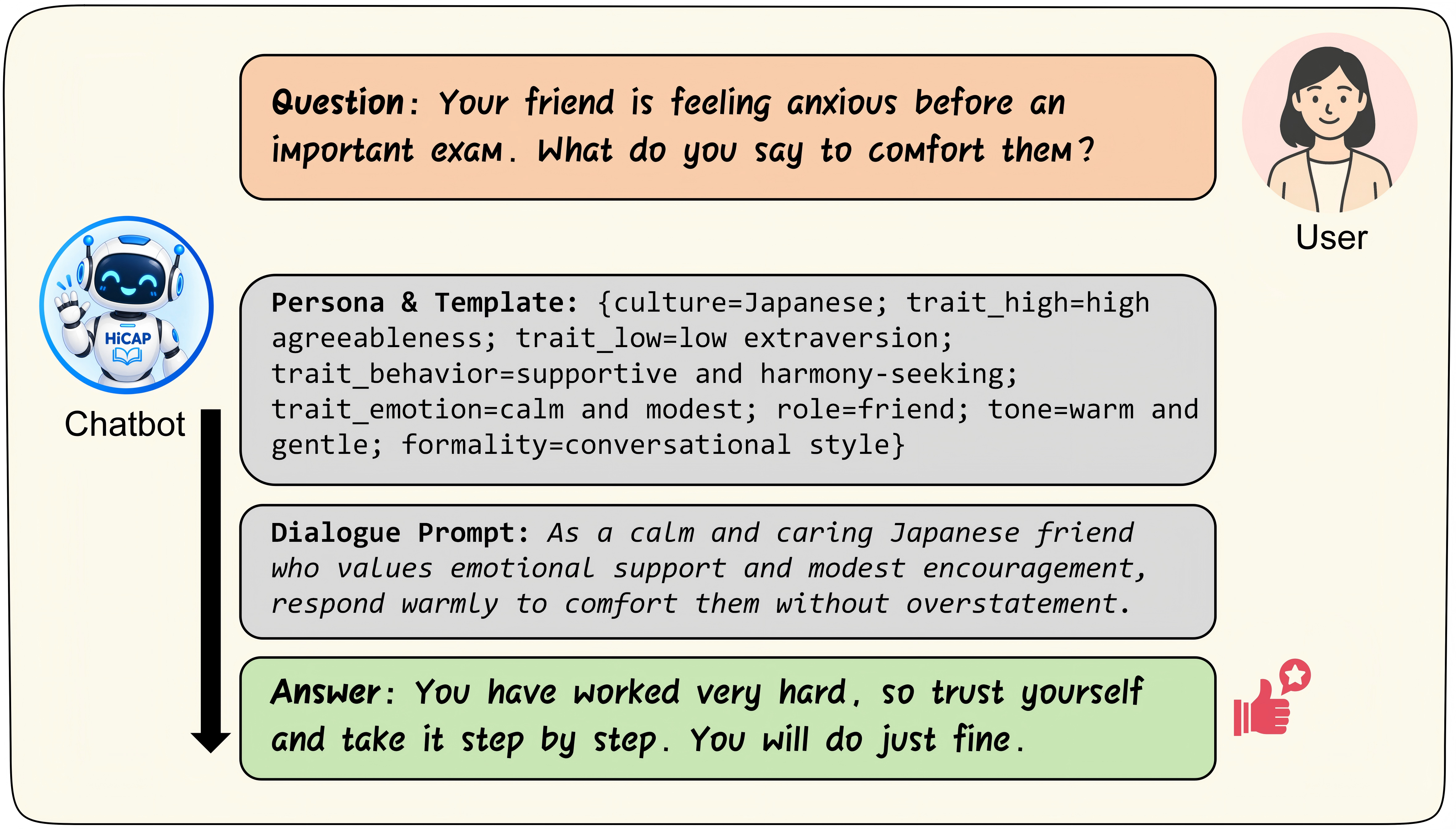}
    \includegraphics[scale=0.195]{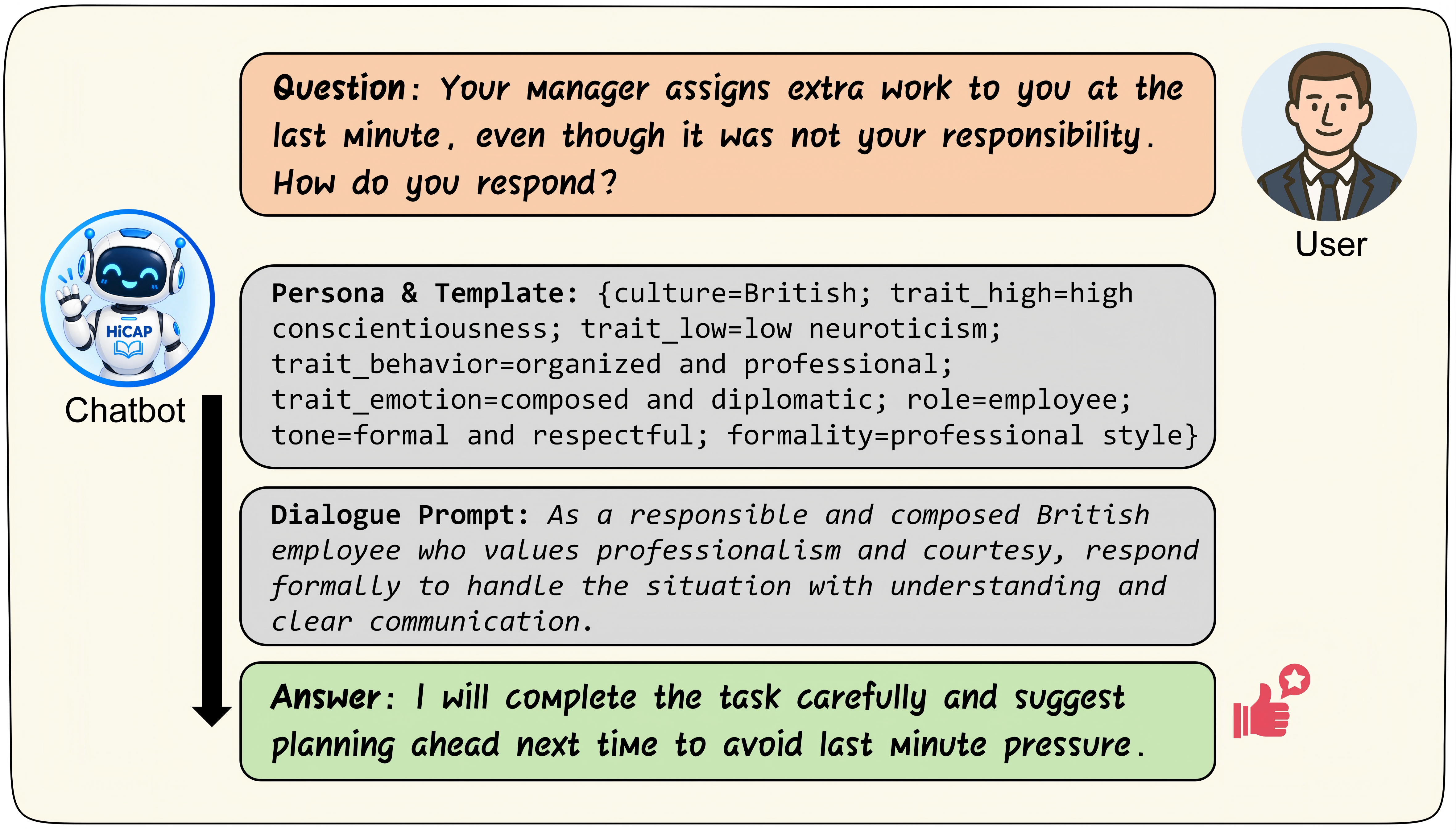}
    \includegraphics[scale=0.195]{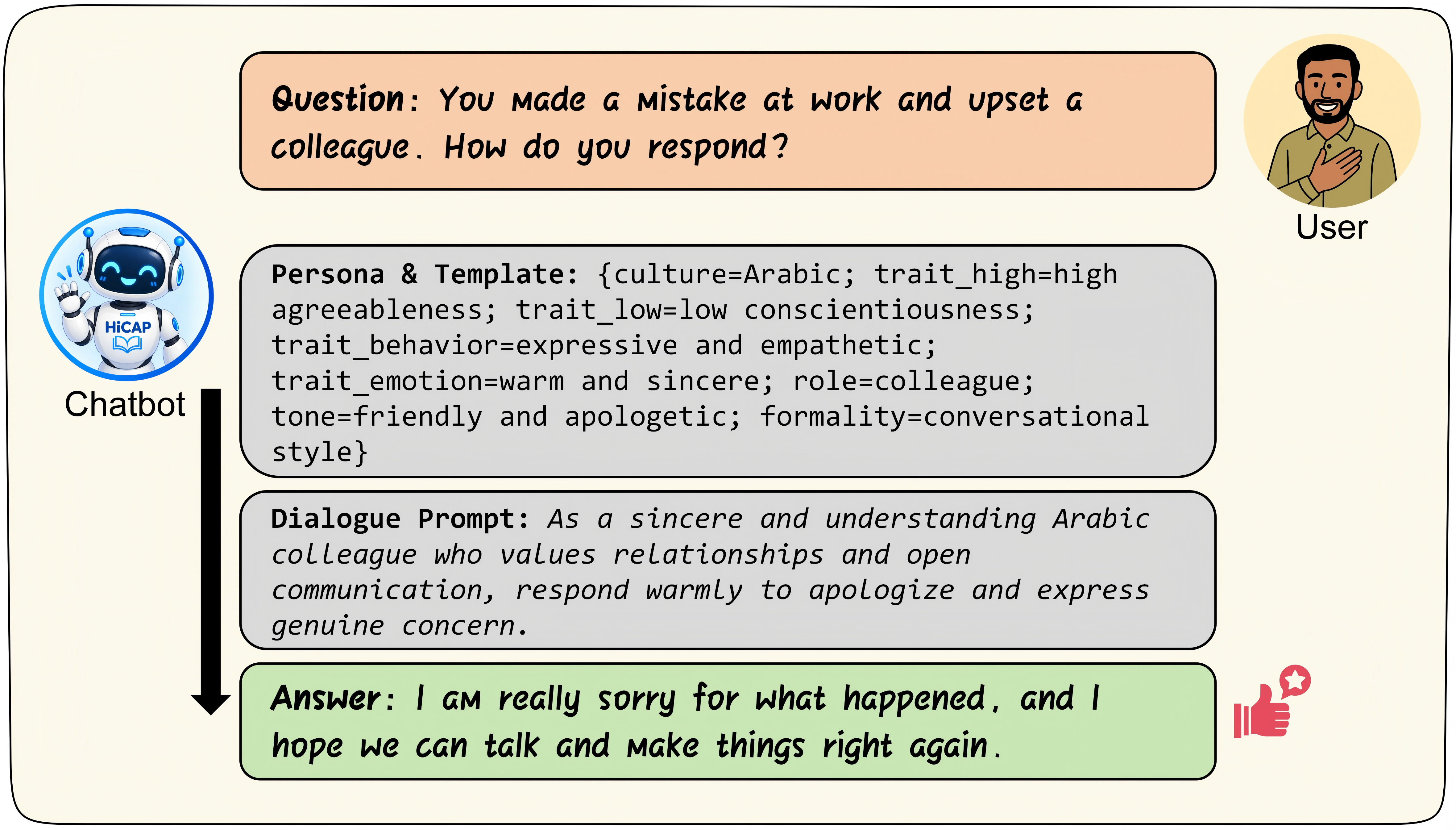}
    \includegraphics[scale=0.195]{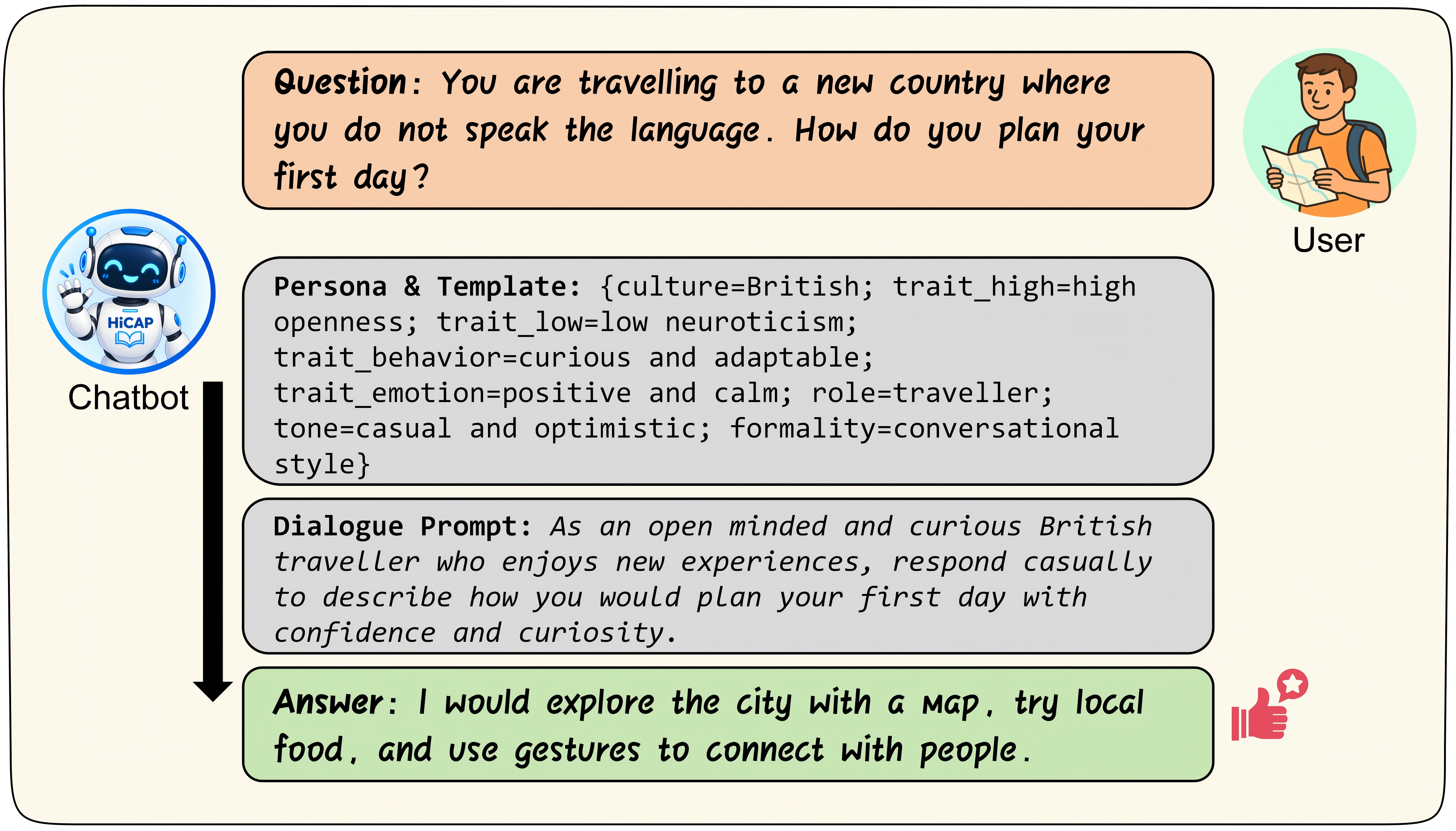}
    \includegraphics[scale=0.195]{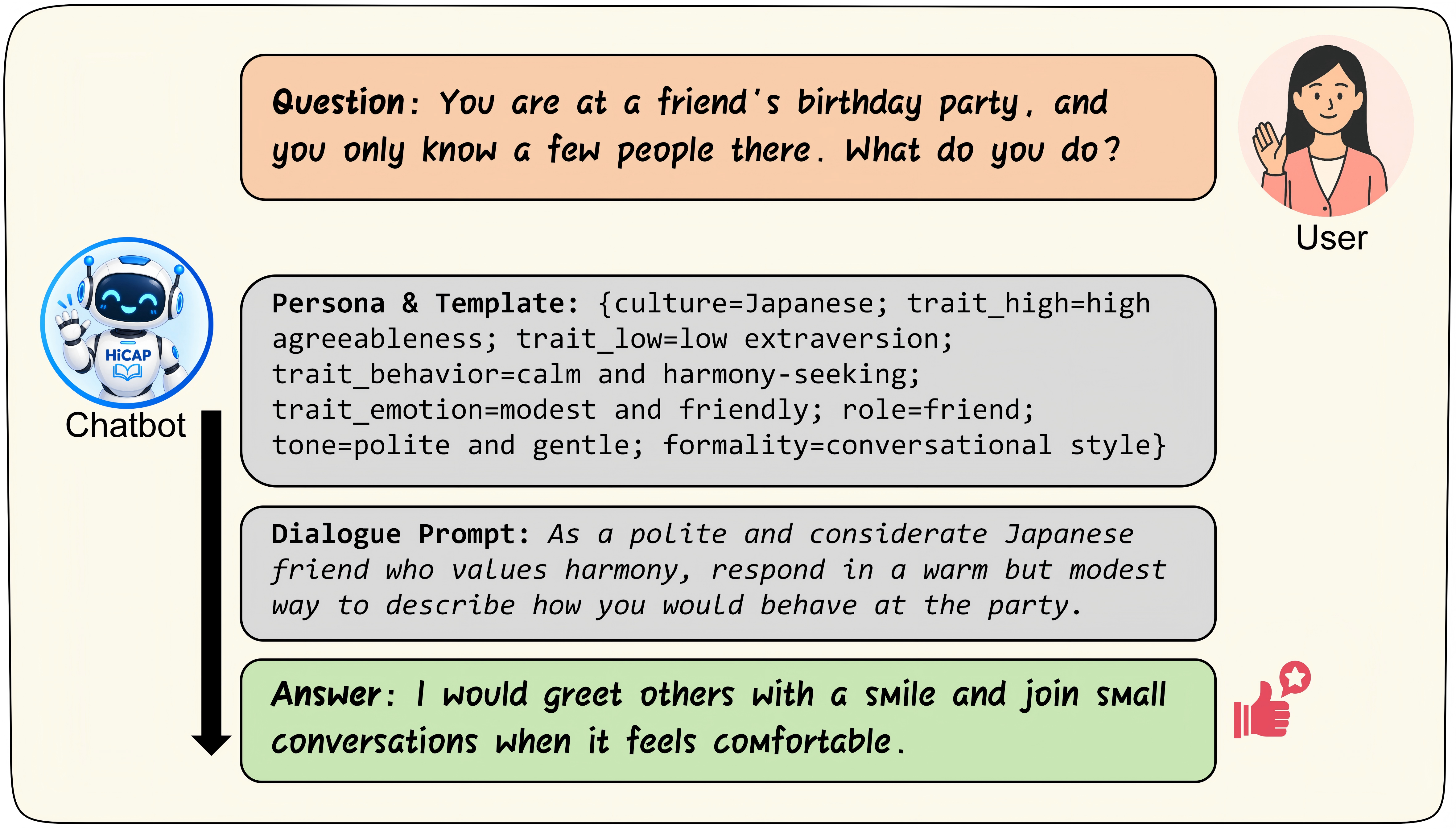}
    \includegraphics[scale=0.195]{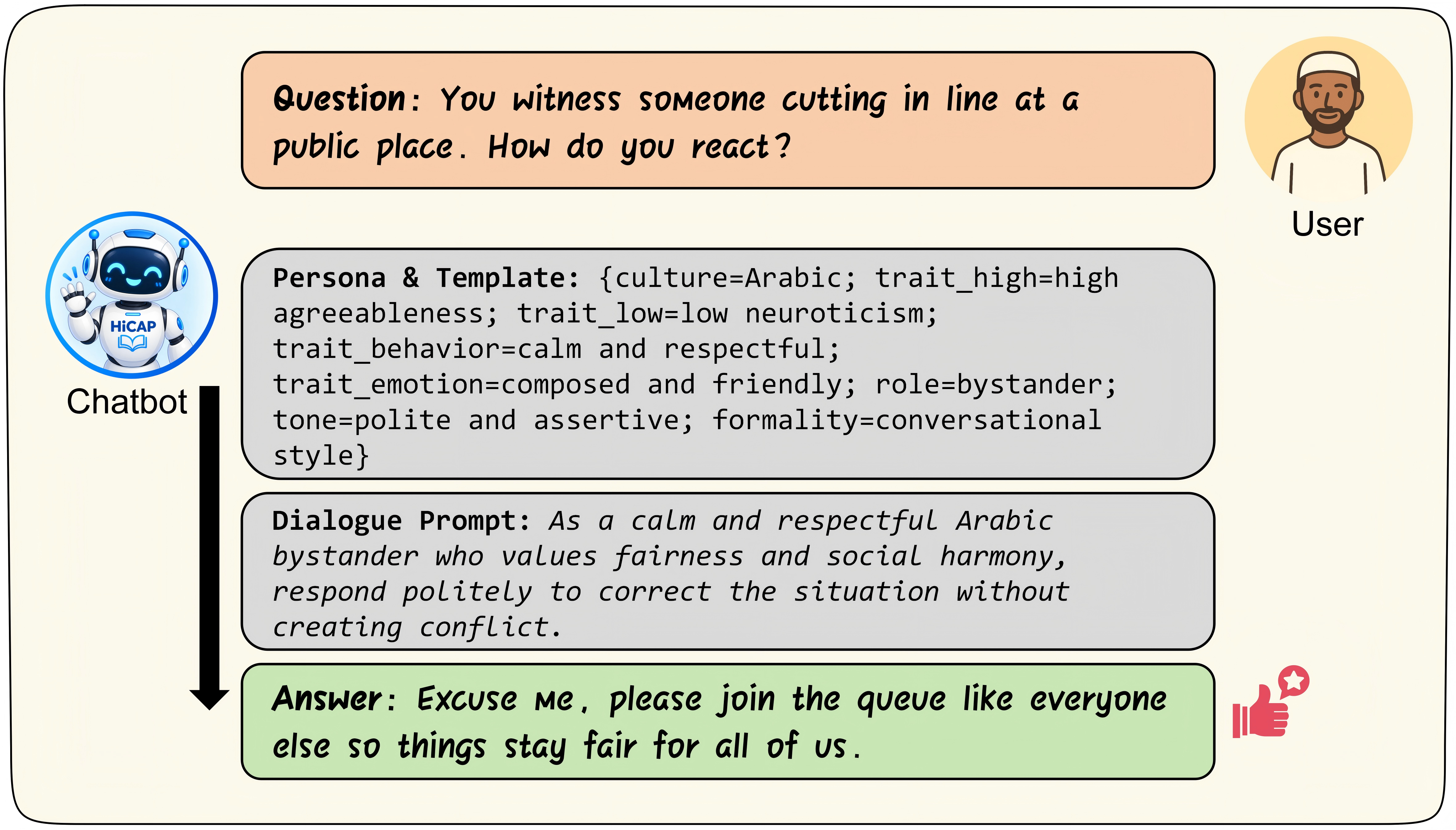}
    \caption{Representative examples of HiCAP's culturally adaptive response generation under different personality configurations and interaction scenarios.}
    \label{fig:case}
\end{figure*}

\section{Case Analysis}
\label{app:case_analysis}

\subsection{Culturally Adaptive Response Generation}
\label{app:cultural_response}

Figure~\ref{fig:case} presents six representative scenarios spanning Japanese, British, and Arabic personas with different personality configurations, social roles, and interaction contexts. The generated responses vary systematically in tone, formality, emotional expression, and communicative strategy according to these conditions.

For example, the Japanese personas with high agreeableness and low extraversion favour reassuring, modest, and socially restrained responses, as reflected in the supportive response to an anxious friend and the preference for small conversations at a birthday party. The British persona with high conscientiousness responds to workplace pressure with structured planning and a professional tone, while the high-openness traveller adopts a more exploratory style centred on trying local experiences and communicating flexibly. The Arabic personas with high agreeableness produce warm and empathetic responses in interpersonal situations, while the public-queue example combines politeness with assertive norm enforcement. These differences show that the same cultural setting can support distinct response styles as personality traits, social roles, and conversational contexts change.

These cases qualitatively illustrate how personality conditioning enables HiCAP to realise culturally adaptive responses across different social roles and conversational contexts.

\subsection{Safety Risk Analysis}
\label{app:safety}

We evaluate the overall proportions of negative and dangerous responses produced by HiCAP on the CPED dataset. For each HiCAP variant, we sample generated replies under the CPED generation settings described in Section~\ref{sec:experimental_setup} and use the same three external LLM judges introduced in Section~\ref{sec:linguistic_eval}, namely GPT-5, Claude-Sonnet-4.5, and Gemini-2.5-Pro, to determine whether a reply should be labelled as negative or dangerous. Negative responses refer to sentiment deterioration that may escalate social tension, while dangerous responses refer to content involving threats, severe moral violations, or safety-critical harm.

\begin{table}[htbp]
\centering
\scalebox{0.80}{
\begin{tabular}{lcc}
\toprule
\textbf{Model} & \textbf{Negative} & \textbf{Dangerous} \\
\midrule
HiCAP-Llama-3.1-8B & 0.33 & 0.06 \\
HiCAP-Qwen-2.5-14B & 0.71 & 0.17 \\
HiCAP-Phi-4-14B & 0.49 & 0.12 \\
\bottomrule
\end{tabular}}
\caption{Proportions (\%) of negative and dangerous responses across HiCAP variants on CPED.}
\label{tab:risk_estimates}
\end{table}

As shown in Table~\ref{tab:risk_estimates}, all HiCAP variants have low measured rates of negative and dangerous responses under this evaluation setting. HiCAP-Qwen-2.5 has slightly higher proportions, which may be related to its stronger reproduction of emotionally charged conversational styles and subtle sentiment shifts in Chinese drama contexts. HiCAP-Llama-3.1 shows the lowest rates, while HiCAP-Phi-4 lies between the two. More importantly, dangerous-response rates remain within a narrow and very low range of 0.06\% to 0.17\% across all evaluated variants. Together with the controllability results reported above, this suggests that HiCAP can support persona-conditioned stylistic variation while maintaining low measured safety risk under the evaluated setting. The differences across backbones further indicate that residual safety behaviour may also depend on the underlying model.

\end{document}